%% file: paper.tex
\documentclass[conference]{IEEEtran}
\IEEEoverridecommandlockouts 
\usepackage{cite}
\usepackage{amsmath,amssymb,amsfonts}
\usepackage[linesnumbered]{algorithm2e}
\usepackage{algpseudocode}
\usepackage{algcompatible}
\usepackage{graphicx}
\usepackage{textcomp}
\usepackage{xcolor}
\usepackage{hyperref}
\usepackage{subcaption}
\usepackage{verbatim}
\usepackage{mathtools}
\usepackage{subfiles} % Best loaded last in the preamble

\def\BibTeX{{\rm B\kern-.05em{\sc i\kern-.025em b}\kern-.08em
    T\kern-.1667em\lower.7ex\hbox{E}\kern-.125emX}}
\begin{document}

    \title{Fast Exploration of the Impact of Precision Reduction on Spiking Neural Networks}

\author{\IEEEauthorblockN{Sepide Saeedi\textsuperscript{1}, Alessio Carpegna\textsuperscript{1}, Alessandro Savino\textsuperscript{1}, and Stefano Di Carlo\textsuperscript{1}}
\IEEEauthorblockA{\textsuperscript{1}\textit{Control and Computer Eng. Dep., Politecnico di Torino}
Torino, Italy\\ \{sepide.saeedi, alessio.carpegna, alessandro.savino, stefano.dicarlo\}@polito.it}
}

\maketitle

\begin{abstract}
Approximate Computing (AxC) techniques trade off the computation accuracy for performance, energy, and area reduction gains. The trade-off is particularly convenient when the applications are intrinsically tolerant to some accuracy loss, as in the Spiking Neural Networks (SNNs) case. SNNs are a practical choice when the target hardware reaches the edge of computing, but this requires some area minimization strategies. In this work, we employ an Interval Arithmetic (IA) model to develop an exploration methodology that takes advantage of the capability of such a model to propagate the approximation error to detect when the approximation exceeds tolerable limits by the application. 
Experimental results confirm the capability of reducing the exploration time significantly, providing the chance to reduce the network parameters' size further and with more fine-grained results.
\end{abstract}


\subfile{sections/introduction}
\subfile{sections/methods}

\subfile{sections/experimental_results}

\subfile{sections/conclusion}

\section*{Acknowledgment}
This work has received funding from the APROPOS project in the European Union’s Horizon 2020 research and innovation programme under the Marie Skłodowska-Curie grant agreement No 956090.\textbf{}

\bibliographystyle{IEEEtran}
\bibliography{bibliography/IEEEabrv.bib, bibliography/biblio.bib}

\end{document}

%% file: sections/introduction.tex
\section{Introduction}
\label{sec:introduction}

Spiking Neural Networks (SNNs) are Artificial Neural Networks (ANN) that mimic human brain functionality more closely than other types of ANNs (e.g., convolutional neural networks). SNNs support information exchange based on binary spikes~\cite{Maass:1997tg} and unsupervised learning with unlabeled data using the Spike-Timing-Dependent Plasticity (STDP)~\cite{Putra:2021tw}. Each neuron in an SNN performs arithmetic operations using weights and thresholds to elaborate the input spikes and produce output spikes. Given the large size of SNN models (i.e., large arrays of registers to store weights and thresholds) and the type of performed operations, SNNs can undoubtedly benefit from employing approximate Computing (AxC) techniques~\cite{Capra:2020uy}, especially when targeting the final deployment at the edge. For instance, when deploying SNNs into FPGA devices, employing AxC techniques can help decrease the storage required for weights and thresholds and the complexity of the arithmetic components~\cite{Ayhan:2018tb, Peng:2018wi, Wang:2019wt}. 

AxC techniques enable a controlled reduction of the computational accuracy to gain in performance, e.g., reductions in power consumption, memory utilization, and execution time~\cite{Stanley-Marbell:2020wb}. This gain is even more interesting when applications, like the SNN, are intrinsically tolerant to some loss in computation accuracy~\cite{Wu:2021tx}. 

When applications introduce AxC techniques, the main issue is selecting what can be approximated and how to do that.  Several methods can be used together. This decision requires a precise evaluation of the impact of each combination of AxC techniques on the final computation accuracy, considering that every application sets a limit for the accuracy degradation so that the results stay within an acceptable range. %Hence, applying a single AxC technique or a combination of the AxC methods to a complex computation, it is necessary to evaluate how much the computation accuracy is degraded and, by losing that amount of computation accuracy, how much gains in performance, e.g., reductions in power consumption, memory utilization, and execution time is obtained. After this evaluation, it is possible to choose the best AxC technique or the best combination of AxC techniques for each application to achieve the minimum computation accuracy loss and maximum gains in performance. 

The two main approaches for evaluating the impact of the application of an AxC technique are running the application several times with different configurations~\cite{Smithson:2016tn, Dupuis:2020tx} or exploiting abstract models to predict the final error of the application \cite{Traiola:2019up, Savino:2021te}. In the first approach, the analysis can be very precise, but the required time increases with the number of available options preventing an exhaustive search of the design space. The second approach trades off the accuracy of the analysis with its computation time to increase the depth of the design space that can be explored. 
%[Maybe a paragraph here that links the ones before and after. It might be helpful to add and remind the reader that the ANNs are intrinsically approximate apps, and there is no single golden accurate answer that we are degrading its accuracy. as long as we reach a specific network accuracy, the computation accuracy loss is acceptable]

This paper proposes an abstract numerical model based on Interval Arithmetics (IA)~\cite{IAstandard} to explore the impact of applying approximate computing techniques to SNNs. The model provides an abstract representation of the SNN computation flow representing how errors introduced by approximations propagate from the inputs to the outputs.  
Among the different approximation techniques, the proposed model focuses on data reduction in fixed-point quantization of weights and thresholds to reduce the final size of the SNN model. Some previous works \cite{Okada:2012vd, Turner:2021vd} proposed using interval arithmetic concepts to analyze neural network computations. The main drawback of these approaches is that they focus on data propagation without providing a way to track the introduced error. This paper proposes a computation flow representation to model the approximation error range, which offers further tuning optimization opportunities. Moreover, the paper presents the concept of watchers during the analysis. Watchers are specific observation points that can be placed in critical points of the computation flow to monitor violations of the acceptable accuracy of the application and help drive the design of space exploration, thus saving time. 

Experiments were performed on a trained SNN designed for deploying FPGA hardware accelerators at the edge~\cite{Carpegna:2022td}. The proposed approach explored the possibility of reducing the network parameters' size to fit the network in small FPGAs used for edge applications. 

The rest of this paper is organized as follows: \autoref{sec:methods} describes the methodology. The experimental results are reported and analyzed in section \autoref{sec:experimental_results}. Eventually, in \autoref{sec:conclusions}, the conclusions and possible future works are provided.

%% file: sections/methods.tex
\section{Methods}
\label{sec:methods}

This section first introduces the spiking neural network architecture considered in the paper and the applied data quantization and precision reduction techniques. Second, it describes the proposed error propagation model and its associated design space exploration methodology.

\subsection{Spike Neuron Model and Network Description}
\label{subsec:spike_nn}

%width=0.9\columnwidth,natwidth=600,natheight=600

\begin{figure}[hbt]
\centering
  \includegraphics[width=0.9\columnwidth]{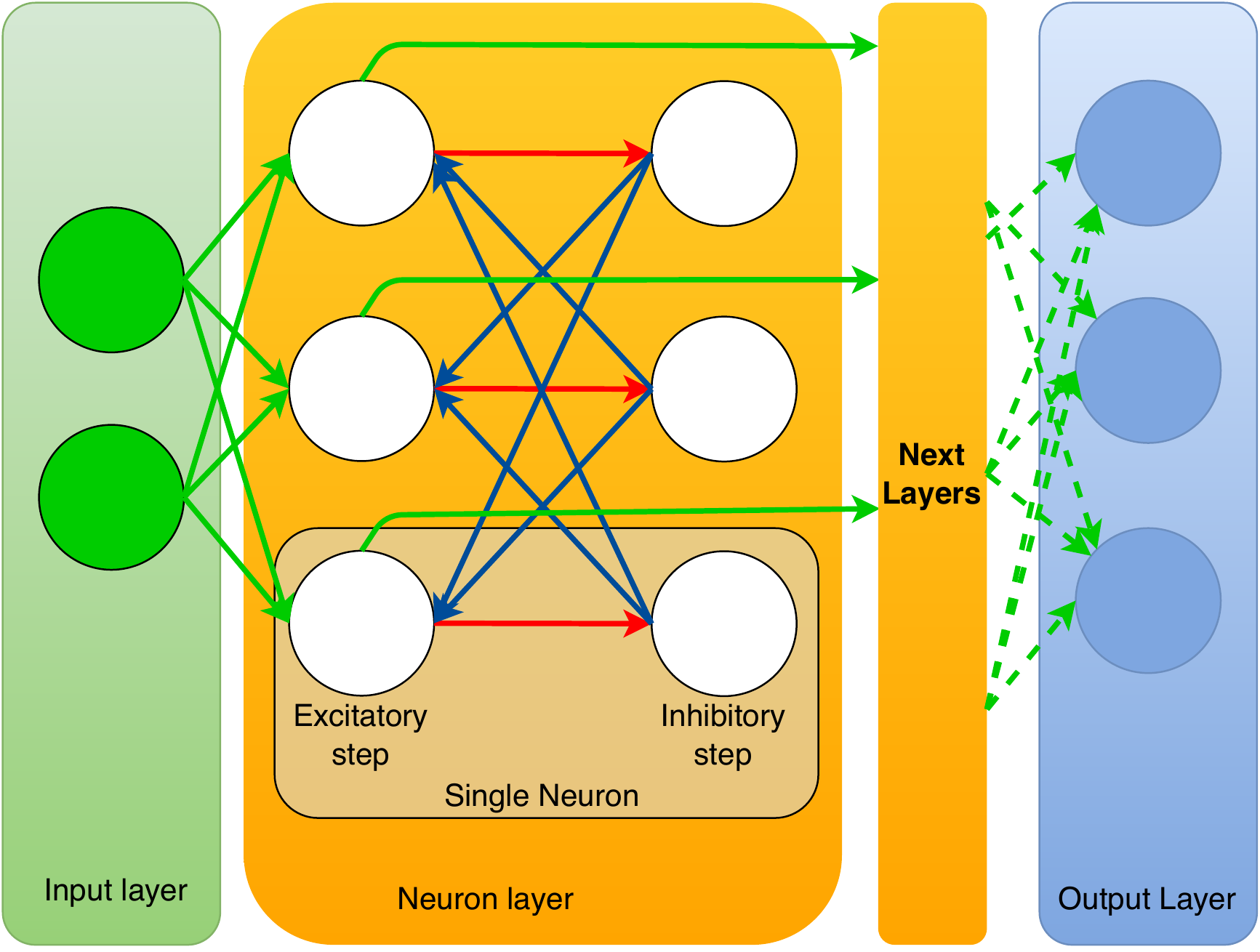}
  \caption{SNN functional model, including a detailed description of one layer}
  \label{fig:snn_one_layer}
\end{figure}

\begin{figure*}[hbt]
  \includegraphics[width=1.00\textwidth]{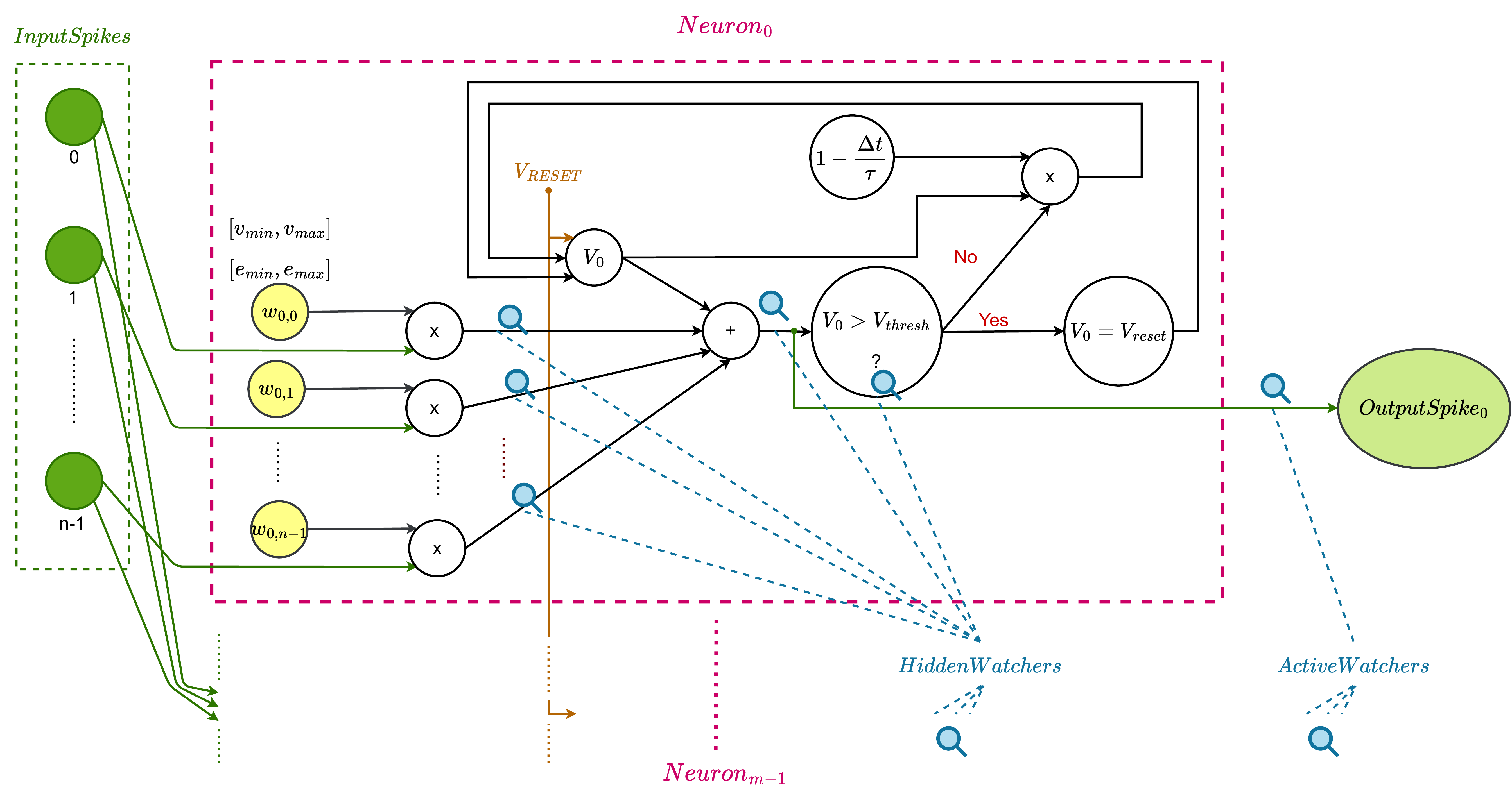}
  \caption{Computation Flow of SNN layer including the watch points used by the exploration algorithm}
  \label{fig:comp_flow}
\end{figure*}

\autoref{fig:snn_one_layer} shows the general organization of an SNN, highlighting the connections in one layer of neurons. Each neuron implements two computational steps: excitatory and inhibitory. Spikes are generated by an input layer transforming data, e.g., pixels of an image, into a sequence of spikes. Spikes are boolean, single-bit information (set to 1 when generated) and enter the neuron through the excitatory step (green arrows from the input layer). This step accumulates a value, called \emph{membrane potential} ($V_0$ in \autoref{fig:comp_flow}), when a spike enters the neuron. An output spike is generated when the \emph{membrane potential} reaches a threshold (the green arrows out the excitatory step). Due to the multi-level structure, those output spikes serve as input for the next layer and as input for the inhibitory step (the red arrow). The inhibitory step models the interaction with all other neurons in the layer, and it acts by decreasing each neuron's \emph{membrane potential} when connected neurons generate spikes. In the end, the final layer is connected to an output layer that receives spikes and, if necessary, transforms them into the network's final result.

Looking closely at the mathematical model (following the flow in \autoref{fig:comp_flow}), each $V_i$ is increased according to a weight ($w_{i,j}$), where $j$ follows the size of the input spikes and $i$ refers to the neuron (e.g., in \autoref{fig:comp_flow}, $i$ is equal to $0$, and $j$ goes from $0$ to $n-1$). As with any ANN, weights are obtained during the training phase. The result is that  $V_i$ is increased by the sum of all weights where spikes are detected in the input connections. This procedure is generalized as shown in \autoref{eq:INxW}. Here, $n$ is the number of input connections of the neuron. $inp_j$ is the input spike and $w_{i,j}$ is the value of each weight for neuron $i$ (as it is shown for the neuron $0$ in \autoref{fig:comp_flow}).

\begin{equation}\label{eq:INxW}
    sum_i = \sum_{j=0}^{n-1} inp_j \cdot w_{i,j}
\end{equation}

When the \emph{membrane potential} reaches its maximum ($V_i > V_{thresh}$), the neuron fires a spike and $V_i$ is reset to a special \emph{reset} value ($V_{reset}$), as depicted in \autoref{fig:comp_flow} following the "Yes" branch. 

To allow comparison of the results without losing generality, this paper applies a set of initial simplifications of the model proposed in \cite{Carpegna:2022tv}. To reduce the number of parameters of the network, a single threshold $V_{thresh}$ is used for each layer. Another optimization is applied to the exponential decay of the \emph{membrane potential}. In general, if no active input spike is detected, $V_i$ should decrease, following an exponential decay approach. To employ a fixed-point number representation, \cite{Carpegna:2022tv} proposed to transform the exponential function into a right shift as depicted in \autoref{fig:comp_flow} by $V_0 \gg exp_{decay}$. Eventually, to comply with the structure proposed in \cite{Carpegna:2022tv}, each neuron is associated with an output spike counter. Using counters is not mandatory and is specific to the SNN exploited in this work. %For more detailed information about the model and implementation, the reader may refer to~\cite{Carpegna:2022td}.

\subsection{Data Quantization and precision reduction}
\label{subsec:dataq_and_precision}

All network parameters are computed during the training phase for each neuron. Since most of the training tools available calculate those values in a floating-point format~\cite{Stimberg:2019wv}, authors in~\cite{Carpegna:2022tv} converted all values into fixed-point numbers. They then explored a precision reduction of the fractional part of 16 bits. The fixed-point format is depicted in \autoref{eq:values}.

\begin{equation}\label{eq:values}
    v = \underbrace{ \sum_{i=16}^{31} b_i \cdot 2^{(i-16)} }_{integer}+ \underbrace{\sum_{i=0}^{15} b_i \cdot 2^{-(16-i)}}_{fractional}
\end{equation}

 A precision reduction of the fractional part aims at reducing the number of bits used for it, ranging from 15 bits to 1 single bit. Since the integer part is not changed, each time the fractional part is reduced, the produced error $\epsilon$ follows \autoref{eq:error}, where $k$ is the number of bits removed from the fractional part and $b_i$ is the value of the original bit removed.

\begin{equation}\label{eq:error}
    \epsilon = \sum_{i=0}^{k} b_i \cdot 2^{-(16-i)}
\end{equation}

For a general value undergoing a precision reduction of $k$, the minimum and the maximum errors introduced are defined in \autoref{eq:error_limits}. Depending on the number of bits ($k$), the maximum error happens when all removed bits ($b_i$) are equal to 1. At the same time, the minimum error occurs when all removed bits ($b_i$) are equal to 0. Hence, the minimum possible error is 0, and the value of the maximum possible error is calculated as in \autoref{eq:error_limits}. 

\begin{equation}\label{eq:error_limits}
    \underbrace{0}_{\forall b_i=0} \leq \epsilon \leq \underbrace{\sum_{i=0}^{k} 2^{-(16-i)}}_{\forall b_i=1} = -\frac{1 - 2^k}{32768}  
\end{equation}

Using \autoref{eq:values}, $v$ can be also expressed as $v - \epsilon$ where $\epsilon$ represents a value subtracted from the fractional part of each number. Such reductions help obtain smaller numbers without compromising the computation accuracy.

\subsection{The Interval Arithmetic Error Propagation Model}
\label{subsec:ia_error_model}

Using the error formulation proposed in \autoref{subsec:dataq_and_precision}, \cite{Saeedi:2022aa} introduced a first error propagation model employing Interval Arithmetics (IA) concepts. The basic concept of IA is defining mathematical operations over intervals instead of single values. An interval is defined as $[v] \equiv [v_{min}, v_{max}]$, where  $v_{min}$ and $v_{max}$ are respectively the minimum and maximum numbers included in the interval. All basic mathematical operations, such as addition and multiplication, can be carried out using Interval Arithmetic concepts~\cite{IAstandard}.

Based on \autoref{eq:error_limits}, the errors can already be represented as a range, defining a generic approximated value as $v - [\epsilon]$. The original value $v$ can also be replaced with an interval. Hence, a generic value in a computation can be represented by a single IA number ($|ian|$) using a pair including two intervals $\{ [v], [\epsilon] \}$. When $[v]$ and $[\epsilon]$ represent a single value instead of a range of values, the two boundaries of the interval must be identical.

To allow the application of the IA algebra presented~\cite{IAstandard} to this new formulation, the proposed model follows the arithmetic formulation of $|ian|$ as $|ian| \simeq [v] - [\epsilon]$. This way, the algebraic operators introduced in~\cite{IAstandard}  can be applied, accepting some approximation in the results. To demonstrate the mathematical adherence of the model, authors in \cite{Saeedi:2022aa} added two considerations:
\begin{enumerate}
	\item the errors are calculated using \autoref{eq:error_limits}, therefore they are strictly monotonic, i.e., $[\epsilon] \equiv [\epsilon_{min}, \epsilon_{max}]$. The same applies to $[v]$.
	\item the relationship between $[v]$ and $[\epsilon]$ is subtractive because the applied approximation technique is a precision reduction approximation. 
\end{enumerate} 

These two considerations are pivotal to defining the mathematical operations necessary to transforming all approximated values, i.e., weights, thresholds, and reset values, into $\{ [v], [\epsilon] \}$ intervals and model the same operations as in the original model. Moreover, this notation allows separating the error from the value itself, which is crucial for exploring the design space effectively as in \autoref{subsec:acceptable_error_range}.

Eventually, let us report how all mathematical operations required by~\autoref{subsec:spike_nn} can be formulated using the proposed IA theory. For this purpose, addition, subtraction, multiplication, right shift, and comparison operands were modeled. As an example, addition and subtraction between two values ($|ian_1|$ and $|ian_2|$) can be defined using the linearity of the two mathematical operations and the monotonic shape of the ranges as in \autoref{eq:add} and \autoref{eq:sub}, resorting to basic interval operations between two intervals. For further details on how the other functions are modeled, the reader may refer to~\cite{Saeedi:2022aa}.

\begin{multline}\label{eq:add}
    |ian_1| + |ian_2| = \{ \underbrace{[v_{1_{min}} + v_{2_{min} }, v_{1_{max}} + v_{2_{max}}]}_{[v]},\\
    				 \underbrace{ [\epsilon_{1_{min}} + \epsilon_{2_{min}}, \epsilon_{1_{max}} + \epsilon_{2_{max}}] }_{[\epsilon]} \}
\end{multline}

\begin{multline}\label{eq:sub}
    |ian_1| - |ian_2| = \{ \underbrace{[v_{1_{min}} - v_{2_{max} }, v_{1_{max}} - v_{2_{min}}]}_{[v]},\\
    				 \underbrace{ [\epsilon_{1_{min}} - \epsilon_{2_{max}}, \epsilon_{1_{max}} - \epsilon_{2_{min}}] }_{[\epsilon]} \}
\end{multline}

\subsection{Monitoring of Acceptable Error Range}
\label{subsec:acceptable_error_range}

Introducing IA to model error propagation is a key tool for implementing a fast design exploration strategy. As indicated in \autoref{fig:comp_flow}, we propose to observe the approximation effects at specific points of the computation flow to monitor the output of the chosen arithmetic operation and modify the precision reduction procedure accordingly. The watchpoints are placed after every mathematical operation and are of two types: (i) \emph{hidden watchers} when they are within the computation flow and evaluate internal computations, and (ii) \emph{active watchers} when they allow observing the output of a neuron. In general, the watchpoints enable us to compare the interval they monitor and the acceptable error range. 

Algorithm~\ref{alg:watcherCheck} depicts the overall evaluation strategy. The exploration begins with setting the maximum precision reduction for all values in the model, i.e., weights and thresholds (lines 5-9). The minimum error is always a 1-bit precision reduction. The maximum error is when a 15-bit precision reduction is applied (line 7). The IA model introduced in \autoref{subsec:ia_error_model} allows for a quick evaluation of the impact of the full range of precision reductions from 1-bit to 16-bit in a single step. 

The watchers are implemented as a class, following an object-oriented programming (OOP) approach, with two methods and few encapsulated data. The first method, \texttt{compareData(...)}, allows evaluating if the error range of the connect $|ian|$ is within an acceptable error range. If not, it sets an internal flag to True. This internal flag is read using a second method, \texttt{watchFired()}. The acceptable error range for all watchers is the mathematical average between the minimum and maximum possible approximation errors as calculated in \autoref{eq:error_limits}. Since the network parameters, such as weights and thresholds, follow a uniform distribution, the approximation-induced errors also follow a uniform distribution. Consequently, we can use the mathematical average between the minimum and maximum approximation errors as the acceptable error.

In \autoref{alg:watcherCheck}, lines 10 to 23 wraps the exploration procedure. The iterative process continues until all model values show no approximation (when $\forall k > 0$ at line 10 is not true anymore) or when no active watcher is fired after a full iteration, i.e., when $explorationDone$ remains False. The first condition tackles the scenario when no approximation is suitable because all precision reductions to explore are set to 0-bit reduction.

Regarding the content of the while loop, line 12 is where the model runs for all inputs. The OOP approach makes it possible to embed the watcher \texttt{compareData()} operation into any computational flow implemented using the $|ian|$ approach.

At the end of the model run, the watchers are checked, as shown in lines 13 to 22. If none of the active watchers, shown in \autoref{fig:comp_flow}, indicate a violation in the error range, the $explorationDone$ flag stays False (as set in line 15), leading to the end of the exploration. Suppose some active watchers indicate a violation in the error range. In that case, a backward analysis is performed, meaning that the hidden watchers are checked from the end to the beginning of the computation flow to find the watchers that fired. Then, the corresponding weight is updated by reducing the number of cut bits by 1. This is possible since the $|ian|$ modeling distinguish between the error and the value.

%% This declares a command \Comment
%% The argument will be surrounded by /* ... */

\SetKwComment{Comment}{/* }{ */}

\begin{algorithm}
\caption{exploration algorithm}\label{alg:watcherCheck}

%\KwData{$n \geq 0$}
%\KwResult{$y = x^n$}
$N := Number\ of\ Neurons$\;
$A := Number\ of\ Active\ Watchers$\;
$H := Number\ of\ Hidden\ Watchers$\;
$explorationDone \gets False$\;

\For{$i=0;\ i<N;\ i++$}{
	\For{$j=0;\ j<H;\ j++$} {
		$k_{best}[i][j] \gets 15$\;
	}
}

%\Comment*[r]{watchers are numbered concerning their place from the first level of the computation flow to the last level. }
%\Comment*[r]{this check should be run in parallel. }

\While{ $explorationDone <> True$ AND $\forall k_{best} > 0$} {
	$explorationDone \gets True$\;
    runExploration()\;
    \For{$i=0;\ i< N;\ i++$}{
        \If{$actWatcher[i].watchFired() == True$}{
        	$explorationDone \gets False$\;
            \For{$j=0;\ j<H;\ j++$}{
                \If{$hidWatcher[i][j].watchFired() == True$}{
                    $k_{best}[i][j]-=1$\;
                }
            } 
        }
    } 
}
\EndWhile
\end{algorithm}

%% file: sections/experimental_results.tex
\section{Experimental Results}
\label{sec:experimental_results}

The proposed approach was applied to optimize the trained SNN network proposed in \cite{Carpegna:2022tv}.  This use case was selected since a simple optimization strategy was used, and it is possible to evaluate if the proposed model can provide benefits. 

The reference SNN was trained with the images from the MNIST dataset~\cite{mnist}. The MNIST dataset consists of 70,000 handwritten images of digits (from 0 to 9), of which 60,000 images are in the training set and 10,000 are in the test set.
For generating the input spikes from the MNIST images, authors in \cite{Carpegna:2022tv} converted the 784 pixels composing each image of size 28x28 pixels into a sequence of spikes employing a random Poisson process~\cite{Heeger:2000uy}, which finally resulted in 3,500 input spikes for each image.

The reference SNN has one layer with 400 neurons. Its primary outputs are the output spike counters, and the secondary outputs are the classification decision by the final decision layer. We focused on the output spike counters to simplify the comparison, avoiding the final classification. The reason is that this work aims to evaluate the quality of the exploration and not the quality of the network itself, including its training approach. Nevertheless, if the model guarantees a correct prediction of the counters layer, the network's classification accuracy remains the same.

The proposed approach was applied considering the behavior of the network when processing 1,000 images of the MNIST test set. We built the IA-based model based on the computational flow in \autoref{fig:comp_flow}. The model was then enhanced to have watchpoints in place. Such model was then used in \autoref{alg:watcherCheck} described previously.
 
\begin{figure}[hbt]
\centering
  \includegraphics[width=0.98\columnwidth]{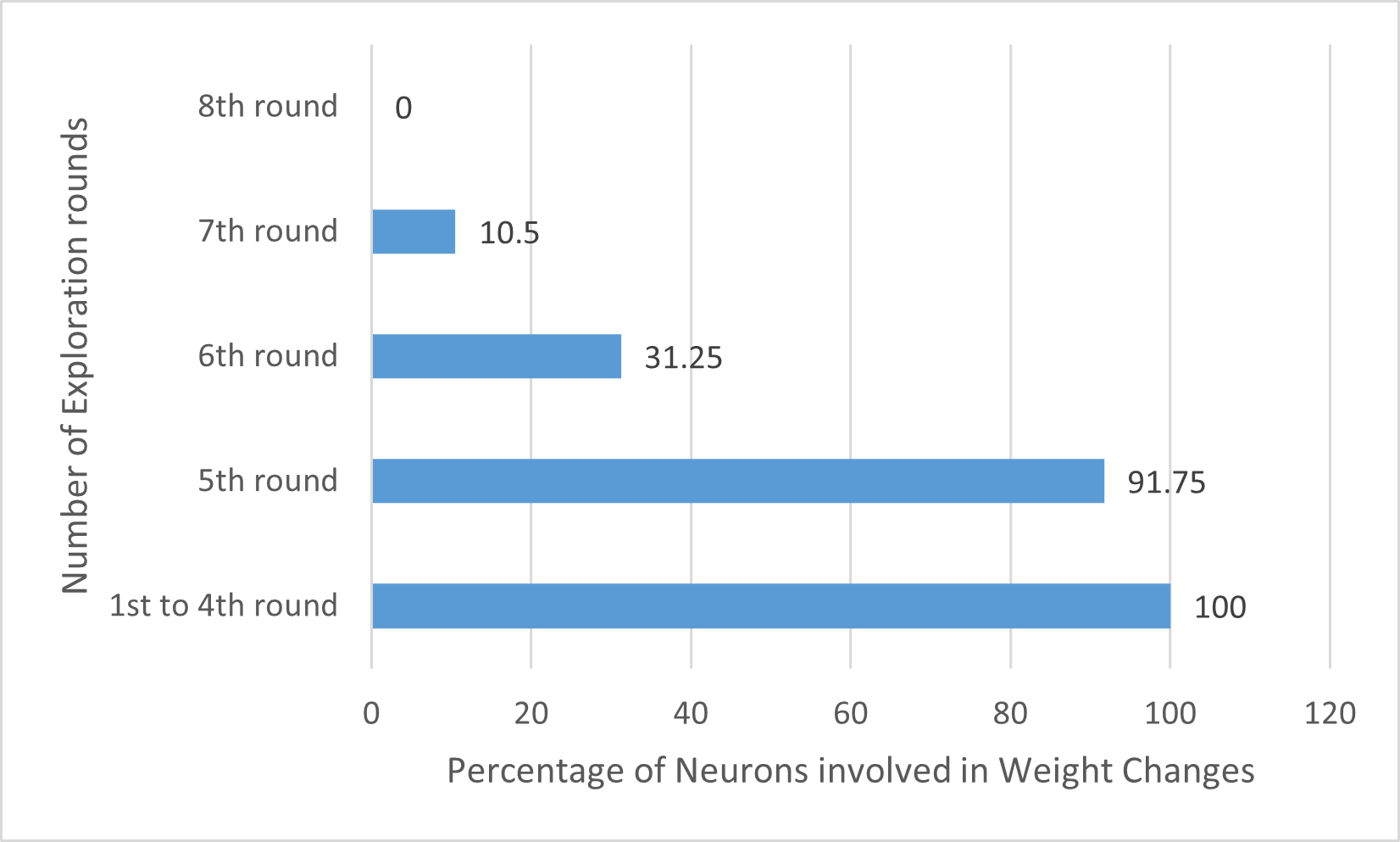}
  \caption{Percentage of neurons involved in changing weights at each round of exploration for one image}
  \label{fig:involved_neurons}
\end{figure}

The exploration always ended before reaching the end condition. On average, the end came after eight rounds of exploration, when no active watchers went off.   \autoref{fig:involved_neurons} shows the number of neurons involved in updating the precision reduction cut at each iteration. In the worst case, the final reduction was down to 10-bits removed from almost all weights as in the reference network~\cite{Carpegna:2022tv}. Nevertheless, the exploration pointed out some weights that were kept with an even higher reduction of 11-bits removed. This is a huge difference from the reference work, where the authors addressed a fast approach to size reduction and imposed the same precision on all values in the model.

At the end of the exploration, to double-check the obtained results, we compared the classification done using the counters computed from the model defined by the exploration with the one generated by the original network in \cite{Carpegna:2022tv}. The network classification accuracy was untouched.

Eventually, let us report some time analysis. The reference system is a laptop computer with 16GB RAM and an Intel Core i7. In the experiments for 1,000 images, eight iterations of exploration required a total exploration time of $539.842$ seconds, and each iteration was carried out on average in $67.480$ seconds. On the contrary, without $|ian|$ and watchpoints tracking, the reference network requires $0.159$ seconds to complete a single image inference. While the raw numbers suggest that the original model is faster, from an exploration perspective, further evaluation is required. Accounting for the model size, each neuron has 784 weights and 400 neurons, so the SNN has 313,600 weights to reduce. Since the precision reduction includes 16 values (from 15 to 0 bits removed), it results in a potential number of combinations of different approximated values equal to $4.18x10^{74}$ as defined by the combination formula in \autoref{eq:weight_comb}. 

Hence, to fairly compare the exploration approach with the reference network, we need to evaluate the time to explore all possible combinations of approximation. In this case, the total required time to run the inference ($0.159$ seconds long) on all $4.18x10^{74}$ combinations is $6.65x10^{73}$, which is unfeasible.  

\begin{equation}\label{eq:weight_comb}
{n \choose r} = \frac{n!}{r!(n-r)!} = {313600 \choose 16} = \frac{313600!}{16!(313584)!}
\end{equation}

At the same time, it is essential to highlight that the authors in the referenced paper did not explore the entire space. This is a crucial aspect when evaluating the results. By working with the parameters of the approximation of every single value in the model, the final configuration is fine-tuned compared with the original one.

%% file: sections/conclusion.tex
\section{Conclusion}
\label{sec:conclusions}

In this paper, we introduced an effective exploration approach for precision reduction approximation of the data defining the model of a Spike Neural Network. The algorithm resorts to a modeling methodology based on Interval Arithmetics that allows tracking the error propagation and accounting for its quantity at each flow operation.

We conducted experiments on a pre-trained network available in the literature that already has a preliminary model reduction. We obtained the same results using the exploration methodology reasonably quickly. Experimental results comparing our model to the original networks confirm that the time spent for the exploration is reduced significantly while providing the opportunity to reduce the precision of weights with more fine-tuned bit precision reductions. 

In future works, we plan to extend the model to other applications with complex computations, including different artificial neural networks, and compare different AxC techniques to further investigate this approach's opportunities. 
 
We also plan to exploit our approach to enhance the exploration with a multi-objective evaluation, including other design parameters, such as power consumption and accuracy.